\documentclass[11pt,a4paper]{article}
\usepackage[hyperref]{acl2018}

\usepackage{times}
\usepackage{latexsym}
\usepackage{url}
\usepackage{graphicx,amsmath,multirow,float}
\usepackage[OT1]{fontenc}
\usepackage{todonotes}
\usepackage{enumitem}

\newcommand{\nop}[1]{}
\newcommand{\mquote}[1]{{``\emph{#1}''}}
\newcommand{\xhdr}[1]{\vspace{1.7mm}\noindent{{\bf #1}}}

\newcommand{\ie}{\emph{i.e.}} 
\newcommand{\eg}{\emph{e.g.}} 

\aclfinalcopy 
\def\aclpaperid{30} 

\title{End-to-End Reinforcement Learning for Automatic Taxonomy Induction}

\author{Yuning Mao$^{1}$, Xiang Ren$^{2}$, Jiaming Shen$^1$, Xiaotao Gu$^1$, Jiawei Han$^1$ \\
$^1$Department of Computer Science, University of Illinois Urbana-Champaign, IL, USA \\
$^2$Department of Computer Science, University of Southern California, CA, USA\\
$^1$\{yuningm2, js2, xiaotao2, hanj\}@illinois.edu $\quad$
  $^2$xiangren@usc.edu $\quad$
}

\begin{document}
\maketitle
\begin{abstract}

We present a novel end-to-end reinforcement learning approach to automatic taxonomy induction from a set of terms. 
While prior methods treat the problem as a two-phase task (\ie, detecting hypernymy pairs followed by organizing these pairs into a tree-structured hierarchy), we argue that such two-phase methods may suffer from error propagation, and cannot effectively optimize metrics that capture the holistic structure of a taxonomy.
In our approach, the representations of term pairs are learned using multiple sources of information and used to determine \textit{which} term to select and \textit{where} to place it on the taxonomy via a policy network.
All components are trained in an end-to-end manner with cumulative rewards, measured by a holistic tree metric over the training taxonomies.
Experiments on two public datasets of different domains show that our approach outperforms prior state-of-the-art taxonomy induction methods up to 19.6\% on ancestor F1.~\footnote{Code and data can be found at \url{https://github.com/morningmoni/TaxoRL}}

\end{abstract}


\section{Introduction}
\begin{figure*}[ht]
    \centering
    \includegraphics[width=16.2cm, height=2cm]{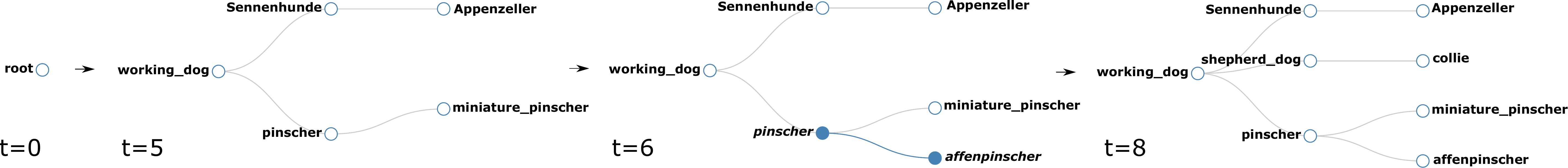}
     \caption{An illustrative example showing the process of taxonomy induction. 
The input vocabulary $V_0$ is \{\mquote{working dog}, \mquote{pinscher}, \mquote{shepherd dog}, ...\}, and the initial taxonomy $T_{0}$ is empty. 
We use a virtual \mquote{root} node to represent $T_{0}$ at $t=0$. 
At time $t=5$, there are 5 terms on the taxonomy $T_{5}$ and 3 terms left to be attached: $V_t$ = \{\mquote{shepherd dog}, \mquote{collie}, \mquote{affenpinscher}\}.
Suppose the term \mquote{affenpinscher} is selected and put under \mquote{pinscher}, then the remaining vocabulary $V_{t+1}$ at next time step becomes \{\mquote{shepherd dog}, \mquote{collie}\}.
Finally, after $|V_0|$ time steps, all the terms are attached to the taxonomy and $V_{|V_0|} = V_{8} = \{ \}$.
A full taxonomy is then constructed from scratch.
}
    \label{fig:toy}
    \vspace*{-.3cm}
\end{figure*}

Many tasks in natural language understanding (\eg, information extraction~\cite{Demeester2016LiftedRI}, question answering~\cite{Yang2017EfficientlyAT}, and textual entailment~\cite{Sammons2012RecognizingTE}) rely on lexical \nop{semantic knowledge} resources in the form of term taxonomies (cf. rightmost column in Fig.~\ref{fig:toy}). 
However, most existing taxonomies, such as WordNet~\cite{miller1995wordnet} and Cyc~\cite{lenat1995cyc}, are manually curated and thus may have limited coverage or become unavailable in some domains and languages. Therefore, recent efforts have been focusing on \textit{automatic taxonomy induction}, which aims to organize a set of terms into a taxonomy based on relevant resources such as text corpora. 

Prior studies on automatic taxonomy induction~\cite{gupta2017taxonomy,camacho2017we} often divide the problem into two sequential subtasks: (1) \textit{hypernymy detection} (\ie, extracting term pairs of ``is-a'' relation); and (2) \textit{hypernymy organization} (\ie, organizing is-a term pairs into a tree-structured hierarchy). 
Methods developed for hypernymy detection either harvest new terms~\cite{yamada2009hypernym,kozareva2010semi} or presume a vocabulary is given and study term semantics~\cite{snow2005learning,fu2014learning,tuan2016learning,shwartz2016improving}.
The hypernymy pairs extracted in the first subtask form a noisy hypernym graph, which is then transformed into a tree-structured taxonomy in the hypernymy organization subtask, using different graph pruning methods including maximum spanning tree (MST)~\cite{bansal2014structured,zhang2016learning}, minimum-cost flow (MCF)~\cite{gupta2017taxonomy} and other pruning heuristics~\cite{kozareva2010semi,velardi2013ontolearn,faralli2015large,panchenko2016taxi}.

However, these two-phase methods encounter two major limitations. First, most of them ignore the taxonomy structure when estimating the probability that a term pair holds the hypernymy relation.
They estimate the probability of different term pairs independently and the learned term pair representations are fixed during hypernymy organization.
In consequence, there is no feedback from the second phase to the first phase and possibly wrong representations cannot be rectified based on the results of hypernymy organization, which causes the error propagation problem. 
Secondly, some methods~\cite{bansal2014structured,zhang2016learning} do explore the taxonomy space by regarding the induction of taxonomy structure as inferring the conditional distribution of edges. 
In other words, they use the product of edge probabilities to represent the taxonomy quality.
However, the edges are treated equally, while in reality, they contribute to the taxonomy differently. 
For example, a high-level edge is likely to be more important than a bottom-out edge because it has much more influence on its descendants.
In addition, these methods cannot \textit{explicitly} capture the holistic taxonomy structure by optimizing global metrics. 

To address the above issues, we propose to jointly conduct hypernymy detection and organization by learning term pair representations and constructing the taxonomy simultaneously.
Since it is infeasible to estimate the quality of all possible taxonomies, we design an end-to-end reinforcement learning (RL) model to combine the two phases.
Specifically, we train an RL agent that employs the term pair representations using multiple sources of information and determines \textit{which} term to select and \textit{where} to place it on the taxonomy via a policy network.
The feedback from hypernymy organization is propagated back to the hypernymy detection phase, based on which the term pair representations are adjusted.
All components are trained in an end-to-end manner with cumulative rewards, measured by a holistic tree metric over the training taxonomies.
The probability of a full taxonomy is no longer a simple aggregated probability of its edges.
Instead, we assess an edge based on how much it can contribute to the whole quality of the taxonomy.

We perform two sets of experiments to evaluate the effectiveness of our proposed approach.
First, we test the end-to-end taxonomy induction performance by comparing our approach with the state-of-the-art two-phase methods, and show that our approach outperforms them significantly on the quality of constructed taxonomies.
Second, we use the same (noisy) hypernym graph as the input of all compared methods, and demonstrate that our RL approach does better hypernymy organization through optimizing metrics that can capture holistic taxonomy structure.

\xhdr{Contributions.}
In summary, we have made the following contributions:
(1) We propose a deep reinforcement learning approach to unify hypernymy detection and organization so as to induct taxonomies in an end-to-end manner.
(2) We design a policy network to incorporate semantic information of term pairs and use cumulative rewards to measure the quality of constructed taxonomies holistically.
(3) Experiments on two public datasets from different domains demonstrate the superior performance of our approach compared with state-of-the-art methods. We also show that our method can effectively reduce error propagation and capture global taxonomy structure.
\nop{
The rest of the paper is organized as follows.
We formalize the problem in Section~\ref{definition}.
We introduce the hypernymy detection model in Section~\ref{hyper}.
We present our RL approach in Section~\ref{rl}. 
We discuss experimental setups and results in Section~\ref{exp}, 
More analysis are conducted in Section~\ref{casestudy}. 
We review related work in Section~\ref{related-work},
and conclude the paper in Section~\ref{conclusion}.
}

\section{Automatic Taxonomy Induction}

\subsection{Problem Definition} \label{definition}
We define a taxonomy $T = (V, R)$ as a tree-structured hierarchy with term set $V$ (\ie, \textit{vocabulary}), and edge set $R$ (which indicates is-a relationship between terms).
A term $v \in V$ can be either a unigram or a multi-word phrase.
The task of \textit{end-to-end taxonomy induction} takes a set of training taxonomies and related resources (\eg, background text corpora) as input, and aims to learn a \textit{model} to construct a full taxonomy $T$ by adding terms from a given vocabulary $V_{0}$ onto an empty hierarchy $T_0$ \textit{one at a time}.
An illustration of the taxonomy induction process is shown in Fig.~\ref{fig:toy}.

\subsection{Modeling Hypernymy Relation} \label{hyper}
Determining which term to select from $V_{0}$ and where to place it on the current hierarchy requires understanding of the semantic relationships between the selected term and all the other terms. We consider multiple sources of information (\ie, resources) for learning hypernymy relation representations of term pairs, including dependency path-based contextual embedding and distributional term embeddings~\cite{shwartz2016improving}.

\xhdr{Path-based Information.}
We extract the shortest dependency paths between each co-occurring term pair from sentences in the given background corpora.
Each path is represented as a sequence of edges that goes from term $x$ to term $y$ in the dependency tree,
and each edge consists of the word lemma, the part-of-speech tag, the dependency label and the edge direction between two contiguous words.
The edge is represented by the concatenation of embeddings of its four components:
\begin{equation*}
\textbf{V}_{e} = [\textbf{V}_{l},, \textbf{V}_{\text{pos}}, \textbf{V}_{\text{dep}}, \textbf{V}_{\text{dir}}].
\end{equation*}

Instead of treating the entire dependency path as a single feature, we encode the sequence of dependency edges $\textbf{V}_{e_1}, \textbf{V}_{e_2}, ..., \textbf{V}_{e_k}$ using an LSTM so that the model can focus on learning from parts of the path that are more informative while ignoring others.
We denote the final output of the LSTM for path $p$ as $\textbf{O}_p$, and use $\mathcal{P}(x, y)$ to represent the set of all dependency paths between term pair $(x, y)$.
A single vector representation of the term pair $(x, y)$ is then computed as $\textbf{P}_{\mathcal{P}(x,y)}$, the weighted average of all its path representations by applying an average pooling:
\begin{equation*}
\textbf{P}_{\mathcal{P}(x,y)} = \frac{\sum_{p \in \mathcal{P}(x, y)} c_{(x,y)}(p) \cdot \textbf{O}_p}{\sum_{p \in \mathcal{P}(x, y)} c_{(x,y)}(p) },
\end{equation*}
where $c_{(x,y)}(p)$ denotes the frequency of path $p$ in $\mathcal{P}(x, y)$.
For those term pairs without dependency paths, we use a randomly initialized \emph{empty path} to represent them as in ~\citet{shwartz2016improving}. 

\xhdr{Distributional Term Embedding.}
The previous path-based features are only applicable when two terms co-occur in a sentence. 
In our experiments, however, we found that only about 17\% of term pairs have sentence-level co-occurrences.\footnote{In comparison, more than 70\% of term pairs have sentence-level co-occurrences in BLESS~\cite{baroni2011we}, a standard hypernymy detection dataset.}
To alleviate the sparse co-occurrence issue, we concatenate the path representation $\textbf{P}_{\mathcal{P}(x,y)}$ with the word embeddings of $x$ and $y$, which capture the distributional semantics of two terms.

\xhdr{Surface String Features.}
In practice, even the embeddings of many terms are missing because the terms in the input vocabulary may be multi-word phrases, proper nouns or named entities,
which are likely not covered by the external pre-trained word embeddings.
To address this issue, we utilize several surface features described in previous studies~\cite{yang2009metric, bansal2014structured, zhang2016learning}.
Specifically, we employ \emph{Capitalization}, \emph{Ends with}, \emph{Contains}, \emph{Suffix match}, \emph{Longest common substring} and \emph{Length difference}.
These features are effective for detecting hypernyms solely based on the term pairs.

\xhdr{Frequency and Generality Features.}
Another feature source that we employ is the hypernym candidates from TAXI\footnote{http://tudarmstadt-lt.github.io/taxi/}~\cite{panchenko2016taxi}.
These hypernym candidates are extracted by lexico-syntactic patterns and may be noisy.
As only term pairs and the co-occurrence frequencies of them  (under specific patterns) are available, we cannot recover the dependency paths between these terms.
Thus, we design two features that are similar to those used in~\cite{panchenko2016taxi,gupta2017taxonomy}.
\footnote{Since the features use additional resource, we wouldn't include them unless otherwise specified.}
\begin{itemize}[leftmargin=*]
  \item \emph{Normalized Frequency Diff.} 
  For a hyponym-hypernym pair $(x_i, x_j)$ where $x_i$ is the hyponym and $x_j$ is the hypernym, its normalized frequency is defined as $\text{freq}_n(x_i, x_j) = \frac{\text{freq}(x_i, x_j)}{\max_k \text{freq}(x_i, x_k)}$, where $\text{freq}(x_i, x_j)$ denotes the raw frequency of $(x_i, x_j)$.
  The final feature score is defined as $\text{freq}_n(x_i, x_j) - \text{freq}_n(x_j, x_i)$, which down-ranks synonyms and co-hyponyms.
  Intuitively, a higher score indicates a higher probability that the term pair holds the hypernymy relation. 
  \item \emph{Generality Diff.} 
  The generality $g(x)$ of a term $x$ is defined as the logarithm of the number of its distinct hyponyms, \ie, $g(x) = log(1 + |\textbf{hypo}|)$, where for any $hypo \in \textbf{hypo}$, $(hypo, x)$ is a hypernym candidate.
  A high $g(x)$ of the term $x$  implies that $x$ is general since it has many distinct hyponyms.
  The generality of a term pair is defined as the difference in generality between $x_j$ and $x_i$: $g(x_j) - g(x_i)$.
  This feature would promote term pairs with the right level of generality and penalize term pairs that are either too general or too specific.
\end{itemize}

The surface, frequency, and generality features are binned and their embeddings are concatenated as a part of the term pair representation.
In summary, the final term pair representation $\textbf{R}_{xy}$ has the following form:
\begin{equation*}
\textbf{R}_{xy} = [\textbf{P}_{\mathcal{P}(x,y)}, \textbf{V}_{w_x}, \textbf{V}_{w_y}, \textbf{V}_{F(x,y)}],
\end{equation*}
where $\textbf{P}_{\mathcal{P}(x,y)}$, $\textbf{V}_{w_x}$, $\textbf{V}_{w_y}$, $\textbf{V}_{F(x,y)}$ denote the path representation, the word embedding of $x$ and $y$, and the feature embeddings, respectively.

Our approach is general and can be flexibly extended to incorporate different feature representation components introduced by other relation extraction models~\cite{zhang2017position,lin2016neural,shwartz2016improving}. We leave in-depth discussion of the design choice of hypernymy relation representation components as future work.


\section{Reinforcement Learning for End-to-End Taxonomy Induction} \label{rl}
We present the reinforcement learning (RL) approach to taxonomy induction in this section.
The RL agent employs the term pair representations described in Section \ref{hyper} as input, and explores how to generate a whole taxonomy by selecting one term at each time step and attaching it to the current taxonomy.
We first describe the environment, including the actions, states, and rewards.
Then, we introduce how to choose actions via a policy network.



\subsection{Actions}
We regard the process of building a taxonomy as making a sequence of actions. 
Specifically, we define that an action $a_t$ at time step $t$ is to (1) select a term $x_1$ from the remaining vocabulary $V_t$; (2) remove $x_1$ from $V_t$, and (3) attach $x_1$ as a hyponym of one term $x_{2}$ that is already on the current taxonomy $T_{t}$.
Therefore, the size of action space at time step $t$ is $|V_t| \times |T_t|$, where $|V_t|$ is the size of the remaining vocabulary $V_t$, and $|T_t|$ is the number of terms on the current taxonomy.
At the beginning of each episode, the remaining vocabulary $V_0$ is equal to the input vocabulary and the taxonomy $T_0$ is empty.
During the taxonomy induction process, the following relations always hold: 
$|V_t| = |V_{t-1}| - 1$, $|T_t| = |T_{t-1}| + 1$, and $|V_t| + |T_t| = |V_0|$.
The episode terminates when all the terms are attached to the taxonomy, which makes the length of one episode equal to $|V_0|$.

A remaining issue is how to select the first term when no terms are on the taxonomy.
One approach that we tried is to add a virtual node as root and consider it as if a real node.
The \emph{root embedding} is randomly initialized and updated with other parameters. 
This approach presumes that all taxonomies share a common root representation and expects to find the real root of a taxonomy via the term pair representations between the virtual root and other terms. 
Another approach that we explored is to postpone the decision of root by initializing $T$ with a random term as current root at the beginning of one episode, and allowing the selection of \emph{new root} by attaching one term as the hypernym of current root.
In this way, it overcomes the lack of prior knowledge when the first term is chosen.
The size of action space then becomes $|A_t| = |V_t| \times |T_t| + |V_t|$, and the length of one episode becomes $|V_0| - 1$.
We compare the performance of the two approaches in Section \ref{exp}.

\subsection{States}
The \emph{state} $\textbf{s}$ at time $t$ comprises the current taxonomy $T_t$ and the remaining vocabulary $V_t$.
At each time step, the environment provides the information of current state, based on which the RL agent takes an action.
Once a term pair $(x_1, x_2)$ is selected, the position of the new term $x_1$ is automatically determined since the other term $x_2$ is already on the taxonomy and we can simply attach $x_1$ by adding an edge between $x_1$ and $x_2$.

\begin{figure*}[t]
    \centering
    \includegraphics[width=15cm,height=6cm]{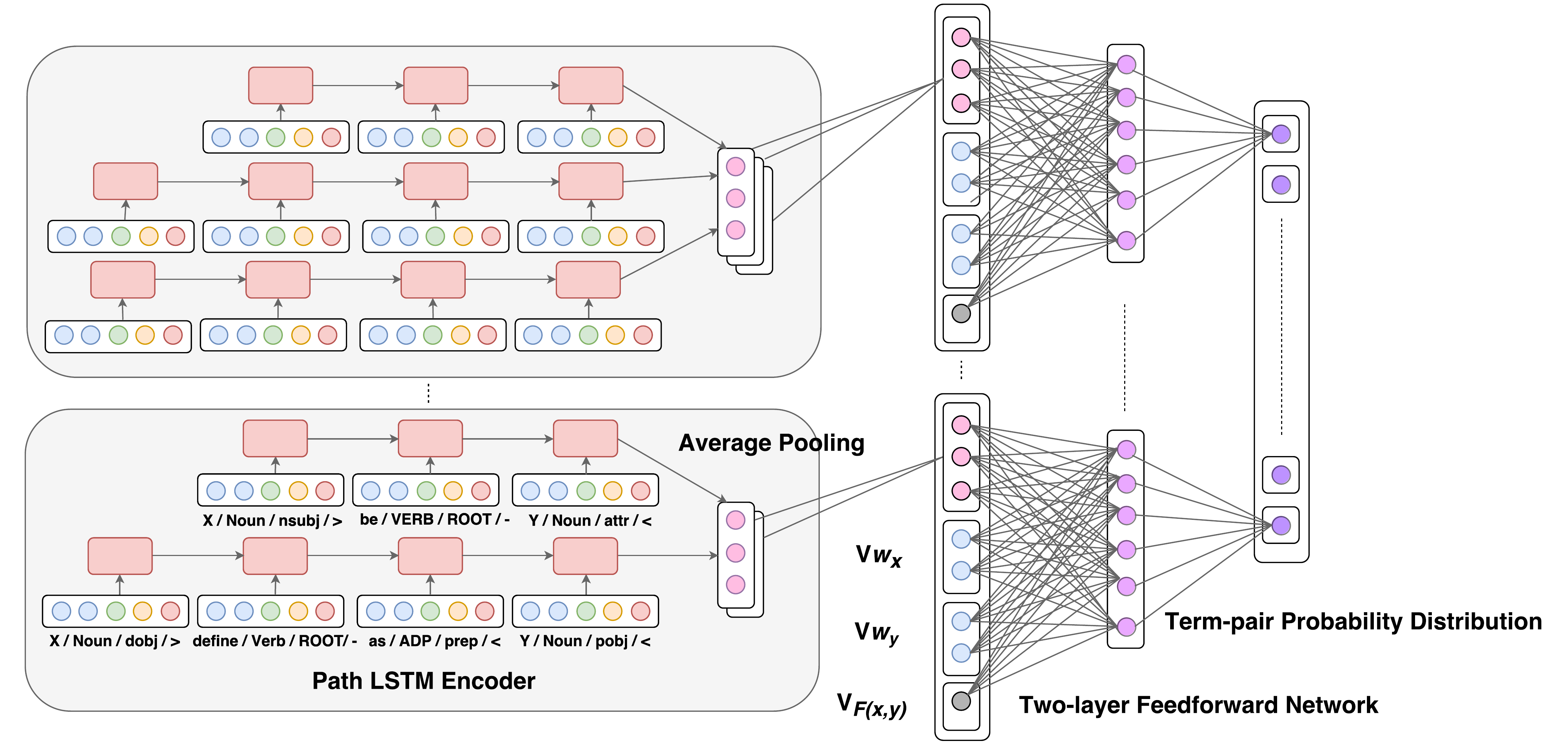}
    \caption[123]{The architecture of the policy network. The dependency paths are encoded and concatenated with word embeddings and feature embeddings, and then fed into a two-layer feed-forward network.}
    \label{fig:net_arch}
    \vspace*{-.35cm}
\end{figure*}

\subsection{Rewards}
The agent takes a scalar \emph{reward} as feedback of its actions to learn its policy.
One obvious reward is to wait until the end of taxonomy induction, and then compare the predicted taxonomy with gold taxonomy.
However, this reward is delayed and difficult to measure individual actions in our scenario.
Instead, we use reward shaping, \ie, giving intermediate rewards at each time step, to accelerate the learning process.
Empirically, we set the reward $r$ at time step $t$ to be the difference of Edge-F1 (defined in Section \ref{metric} and evaluated by comparing the current taxonomy with the gold taxonomy) between current and last time step: $r_t = F1_{e_t} - F1_{e_{t-1}}$.
If current Edge-F1 is better than that at last time step, the reward would be positive, and vice versa.
The cumulative reward from current time step to the end of an episode would cancel the intermediate rewards and thus reflect whether current action improves the overall performance or not.
As a result, the agent would not focus on the selection of current term pair but have a long-term view that takes following actions into account.
For example, suppose there are two actions at the same time step. One action attaches a leaf node to a high-level node, and the other action attaches a non-leaf node to the same high-level node. Both attachments form a wrong edge but the latter one is likely to receive a higher cumulative reward because its following attachments are more likely to be correct.


\subsection{Policy Network}
After we introduce the term pair representations and define the states, actions, and rewards, the problem becomes how to choose an action from the action space, \ie, which term pair ($x_{1}, x_{2}$) should be selected given the current state?
To solve the problem, we parameterize each action $a$ by a policy network $\pi(\textbf{a}\ |\ \textbf{s}; \textbf{W}_{RL})$.
The architecture of our policy network is shown in Fig.~\ref{fig:net_arch}.
For each term pair, its representation is obtained by the path LSTM encoder, the word embeddings of both terms, and the embeddings of features.
By stacking the term pair representations, we can obtain an action matrix $\textbf{A}_t$ with size $(|V_t| \times |T_t|) \times dim(\textbf{R})$, where $(|V_t| \times |T_t|)$ denotes the number of possible actions (term pairs) at time $t$ and $dim(\textbf{R})$ denotes the dimension of term pair representation $\textbf{R}$.
$\textbf{A}_t$ is then fed into a two-layer feed-forward network followed by a softmax layer which outputs the probability distribution of actions.\footnote{We tried to encode induction history by feeding representations of previously selected term pairs into an LSTM, and leveraging the output of the LSTM as history representation (concatenating it with current term pair representations or passing it to a feed-forward network).
However, we didn't observe clear performance change.}
Finally, an action $a_{t}$ is sampled based on the probability distribution of the action space:
\begin{equation*}
\begin{split}
\textbf{H}_t &= \text{ReLU} (\textbf{W}_{RL}^1 \textbf{A}^T_t  + \textbf{b}_{RL}^1), \\
\pi(\textbf{a}\ |\ \textbf{s}; \textbf{W}_{RL}) &= \text{softmax} ( \textbf{W}_{RL}^2 \textbf{H}_t + \textbf{b}_{RL}^2), \\
a_t &\sim \pi(\textbf{a}\ |\ \textbf{s}; \textbf{W}_{RL}).
\end{split} 
\end{equation*}
At the time of inference, instead of sampling an action from the probability distribution, we greedily select the term pair with the highest probability. 

We use \emph{REINFORCE}~\cite{williams1992simple}, one instance of the policy gradient methods as the optimization algorithm.
Specifically, for each episode, the weights of the policy network are updated as follows:
\begin{equation*}
	\textbf{W}_{RL} = \textbf{W}_{RL} + \alpha \sum_{t=1}^T \nabla log \pi(\textbf{a}_t\ |\ \textbf{s}; \textbf{W}_{RL}) \cdot v_t,
\end{equation*}
where $v_i = \sum_{t=i}^T \gamma^{t-i} r_t$ is the culmulative future reward at time $i$ and $\gamma \in [0, 1]$ is a discounting factor of future rewards.

To reduce variance, 10 rollouts for each training sample are run and the rewards are averaged. 
Another common strategy for variance reduction is to use a baseline and give the agent the difference between the real reward and the baseline reward instead of feeding the real reward directly.
We use a moving average of the reward as the baseline for simplicity.

\subsection{Implementation Details}
We use pre-trained GloVe word vectors~\cite{pennington2014glove} with dimensionality 50 as word embeddings.
We limit the maximum number of dependency paths between each term pair to be 200 because some term pairs containing general terms may have too many dependency paths.
We run with different random seeds and hyperparameters and use the validation set to pick the best model.
We use an Adam optimizer with initial learning rate $10^{-3}$.
We set the discounting factor $\gamma$ to 0.4 as it is shown that using a smaller discount factor than defined can be viewed as regularization~\cite{jiang2015dependence}.
Since the parameter updates are performed at the end of each episode, we cache the term pair representations and reuse them when the same term pairs are encountered again in the same episode.
As a result, the proposed approach is very time efficient -- each training epoch takes less than 20 minutes on a single-core CPU using DyNet~\cite{neubig2017dynet}.


\section{Experiments} \label{exp}
We design two experiments to demonstrate the effectiveness of our proposed RL approach for taxonomy induction.
First, we compare our end-to-end approach with two-phase methods and show that our approach yields taxonomies with higher quality through reducing error propagation and optimizing towards holistic metrics.
Second, we conduct a controlled experiment on hypernymy organization, where the same hypernym graph is used as the input of both our approach and the compared methods. We show that our RL method is more effective at hypernymy organization.

\subsection{Experiment Setup}
Here we introduce the details of our two experiments on validating that (1) the proposed approach can effectively reduce error propagation; and (2) our approach yields better taxonomies via optimizing metrics on holistic taxonomy structure.


\xhdr{Performance Study on End-to-End Taxonomy Induction.} 
In the first experiment, we show that our joint learning approach is superior to two-phase methods.
Towards this goal, we compare with TAXI~\cite{panchenko2016taxi}, a typical two-phase approach, two-phase HypeNET, implemented by pairwise hypernymy detection and hypernymy organization using MST, and~\citet{bansal2014structured}.
The dataset we use in this experiment is from~\citet{bansal2014structured}, which is a set of medium-sized full-domain taxonomies consisting of bottom-out full subtrees sampled from WordNet.
Terms in different taxonomies are from various domains such as animals, general concepts, daily necessities.
Each taxonomy is of height four (\ie, 4 nodes from root to leaf) and contains (10, 50] nodes.
The dataset contains 761 non-overlapped taxonomies in total and is partitioned by 70/15/15\% (533/114/114) as training, validation, and test set, respectively.

\xhdr{Testing on Hypernymy Organization.} 
In the second experiment, we show that our approach is better at hypernymy organization by leveraging the global taxonomy structure.
For a fair comparison, we reuse the hypernym graph as in TAXI~\cite{panchenko2016taxi} and SubSeq~\cite{gupta2017taxonomy} so that the inputs of each model are the same.
Specifically, we restrict the action space to be the same as the baselines by considering only term pairs in the hypernym graph, rather than all $|V| \times |T|$ possible term pairs.
As a result, it is possible that at some point no more hypernym candidates can be found but the remaining vocabulary is still not empty.
If the induction terminates at this point, we call it a \emph{partial induction}.
We can also continue the induction by restoring the original action space at this moment so that all the terms in $V$ are eventually attached to the taxonomy.
We call this setting a \emph{full induction}.
In this experiment, we use the English environment and science taxonomies in the SemEval-2016 task 13 (TExEval-2)~\cite{bordea2016semeval}.
Each taxonomy is composed of hundreds of terms, which is much larger than the WordNet taxonomies.
The taxonomies are aggregated from existing resources such as WordNet, Eurovoc\footnote{http://eurovoc.europa.eu/drupal/}, and the Wikipedia Bitaxonomy~\cite{flati2014two}.
Since this dataset provides no training data, we train our model using the WordNet dataset in the first experiment.
To avoid possible overlap between these two sources, we exclude those taxonomies constructed from WordNet.


In both experiments, we combine three public corpora -- the latest Wikipedia dump, the UMBC web-based corpus~\cite{han2013umbc_ebiquity} and the One Billion Word Language Modeling Benchmark~\cite{chelba2013one}.
Only sentences where term pairs co-occur are reserved, which results in a corpus with size 2.6 GB for the WordNet dataset and 810 MB for the TExEval-2 dataset. 
Dependency paths between term pairs are extracted from the corpus via spaCy\footnote{https://spacy.io/}.

\subsection{Evaluation Metrics} \label{metric}
 
\xhdr{Ancestor-F1.} It compares the ancestors (``is-a'' pairs) on the predicted taxonomy with those on the gold taxonomy.
We use $P_{a}$, $R_{a}$, $F1_a$ to denote the precision, recall, and F1-score, respectively:
\begin{equation*}
\begin{split}
P_{a} = \frac{| \text{is-a}_{\text{sys}} \wedge \text{is-a}_{\text{gold}} |} {| \text{is-a}_{\text{sys}} |},
R_{a} &= \frac{| \text{is-a}_{\text{sys}} \wedge \text{is-a}_{\text{gold}} |} {| \text{is-a}_{\text{gold}} |}.\\
\end{split}
\end{equation*}
\xhdr{Edge-F1.} It is more strict than \emph{Ancestor-F1} since it only compares predicted edges with gold edges. 
Similarly, we denote edge-based metrics as $P_e$, $R_e$, and $F1_e$, respectively.
Note that $P_e = R_e = F1_e$ if the number of predicted edges is the same as gold edges. 


\begin{table}[t]
    \centering
    \footnotesize
    \setlength\tabcolsep{3pt}
    \scalebox{0.97}{
    {\begin{tabular}{|c|c|c|c||c|c|c|c|}
        \hline
        Model  & $P_a$ & $R_a$ & $F1_a$ & $P_e$ & $R_e$ & $F1_e$ \\
        \hline
         TAXI & 66.1 & 13.9 & 23.0 & 54.8 & 18.0 & 27.1 \\
        \hline
        HypeNET & 32.8 & 26.7 & 29.4 & 26.1 & 17.2 & 20.7 \\
        \hline
        HypeNET+MST & 33.7 & 41.1 & 37.0 & 29.2 & 29.2 & 29.2 \\
        \hline
        TaxoRL (RE) & 35.8 & 47.4 & 40.8 & 35.4 & 35.4 & 35.4 \\
        \hline
        TaxoRL (NR) & 41.3 & 49.2 & \textbf{44.9} & 35.6 & 35.6 & \textbf{35.6} \\
        \hline
        \hline
       \citet{bansal2014structured} & 48.0 & 55.2 & 51.4 & - & - & - \\
        \hline
        TaxoRL (NR) + FG & 52.9 & 58.6 & \textbf{55.6} & 43.8 & 43.8 & \textbf{43.8} \\
        \hline
        \end{tabular}}
        }
    \caption{Results of the end-to-end taxonomy induction experiment. Our approach significantly outperforms two-phase methods~\cite{panchenko2016taxi,shwartz2016improving,bansal2014structured}. \citet{bansal2014structured} and TaxoRL (NR) + FG are listed separately because they use extra resources.} 
    \label{table_wn}
    \vspace*{-.4cm}
\end{table}

\subsection{Results} 
\xhdr{Comparison on End-to-End Taxonomy Induction.} 
Table~\ref{table_wn} shows the results of the first experiment. 
HypeNET~\cite{shwartz2016improving} uses additional surface features described in Section~\ref{hyper}.
HypeNET+MST extends HypeNET by first constructing a hypernym graph using HypeNET's output as weights of edges and then finding the MST~\cite{chu1965shortest} of this graph. 
TaxoRL (RE) denotes our RL approach which assumes a common \emph{Root Embedding}, and TaxoRL (NR) denotes its variant that allows a \emph{New Root} to be added.

We can see that TAXI has the lowest $F1_a$ while HypeNET performs the worst in $F1_e$.
Both TAXI and HypeNET's $F1_a$ and $F1_e$ are lower than 30.
HypeNET+MST outperforms HypeNET in both $F1_a$ and $F1_e$, because it considers the global taxonomy structure, although the two phases are performed independently.
TaxoRL (RE) uses exactly the same input as HypeNET+MST and yet achieves significantly better performance, which demonstrates the superiority of combining the phases of hypernymy detection and hypernymy organization.
Also, we found that presuming a shared root embedding for all taxonomies can be inappropriate if they are from different domains, which explains why TaxoRL (NR) performs better than TaxoRL (RE).
Finally, after we add the frequency and generality features (TaxoRL (NR) + FG), our approach outperforms~\citet{bansal2014structured}, even if a much smaller corpus is used.\footnote{\citet{bansal2014structured} use an unavailable resource~\cite{brants2006web} which contains one trillion tokens while our public corpus contains several billion tokens. The frequency and generality features are sparse because the vocabulary that TAXI (in the TExEval-2 competition) used for focused crawling and hypernymy detection was different.}

\begin{table}[t]
\begin{footnotesize}
    \centering
    \setlength\tabcolsep{3pt}
    \scalebox{0.97}{
    \begin{tabular}{|c|c|c|c|c||c|c|c|c|}
        \hline
       \textbf{} & Model  & $P_a$ & $R_a$ & $F1_a$ & $P_e$ & $R_e$ & $F1_e$ \\
        \hline
        \multirow{5}{*}{Env} &TAXI (DAG) & 50.1 & 32.7 & 39.6 & 33.8  & 26.8 & 29.9 \\
                                                                                            &TAXI (tree) & \textbf{67.5} & 30.8 & 42.3 & \textbf{41.1}  & 23.1 & 29.6 \\
                                                                                            &SubSeq & - & - & - & -  & - & 22.4 \\
                                                                                            &TaxoRL (Partial) & 51.6 & 36.4 & 42.7 & 37.5  & 24.2 & 29.4 \\
                                                                                            &TaxoRL (Full) & 47.2 & \textbf{54.6} & \textbf{50.6} & 32.3  & \textbf{32.3} & \textbf{32.3} \\
        \hline
        
                \multirow{5}{*}{Sci} &TAXI (DAG) & 61.6 & 41.7 & 49.7 & 38.8  & 34.8 & 36.7 \\
                                                                  &TAXI (tree) & 76.8 & 38.3 & 51.1 & 44.8  & 28.8 & 35.1 \\
                                                                  &SubSeq & - & - & - & -  & - & 39.9 \\
                                                                  &TaxoRL (Partial) & \textbf{84.6} & 34.4 & 48.9 & \textbf{56.9}  & 33.0 & \textbf{41.8} \\
                                                                  &TaxoRL (Full) & 68.3 & \textbf{52.9} & \textbf{59.6} & 37.9  & \textbf{37.9} & 37.9 \\
        \hline
        \end{tabular}
    }
    \caption{Results of the hypernymy organization experiment. Our approach outperforms~\citet{panchenko2016taxi,gupta2017taxonomy} when the same hypernym graph is used as input. The precision of partial induction in both metrics is high. The precision of full induction is relatively lower but its recall is much higher.} 
    \label{table_semeval}
\end{footnotesize}
\vspace*{-.4cm}
\end{table}

\xhdr{Analysis on Hypernymy Organization.} 
Table~\ref{table_semeval} lists the results of the second experiment.
TAXI (DAG)~\cite{panchenko2016taxi} denotes TAXI's original performance on the TExEval-2 dataset.\footnote{alt.qcri.org/semeval2016/task13/index.php?id=evaluation}
Since we don't allow DAG in our setting, we convert its results to trees (denoted by TAXI (tree)) by only keeping the first parent of each node.
SubSeq~\cite{gupta2017taxonomy} also reuses TAXI's hypernym candidates.
TaxoRL (Partial) and TaxoRL (Full) denotes partial induction and full induction, respectively.
Our joint RL approach outperforms baselines in both domains substantially.
TaxoRL (Partial) achieves higher precision in both ancestor-based and edge-based metrics but has relatively lower recall since it discards some terms.
In addition, it achieves the best $F1_e$ in science domain.
TaxoRL (Full) has the highest recall in both domains and metrics, with the compromise of lower precision.
Overall, TaxoRL (Full) performs the best in both domains in terms of $F1_a$ and achieves best $F1_e$ in environment domain.


\section{Ablation Analysis and Case Study} \label{casestudy}
In this section, we conduct ablation analysis and present a concrete case for better interpreting our model and experimental results.

Table~\ref{table_ab} shows the ablation study of TaxoRL (NR) on the WordNet dataset.
As one may find, different types of features are complementary to each other.
Combining distributional and path-based features performs better than using either of them alone~\cite{shwartz2016improving}. Adding surface features helps model string-level statistics that are hard to capture by distributional or path-based features. Significant improvement is observed when more data is used, meaning that standard corpora (such as Wikipedia) might not be enough for complicated taxonomies like WordNet.

\begin{figure}[!t]
    \centering
    \includegraphics[width=8cm]{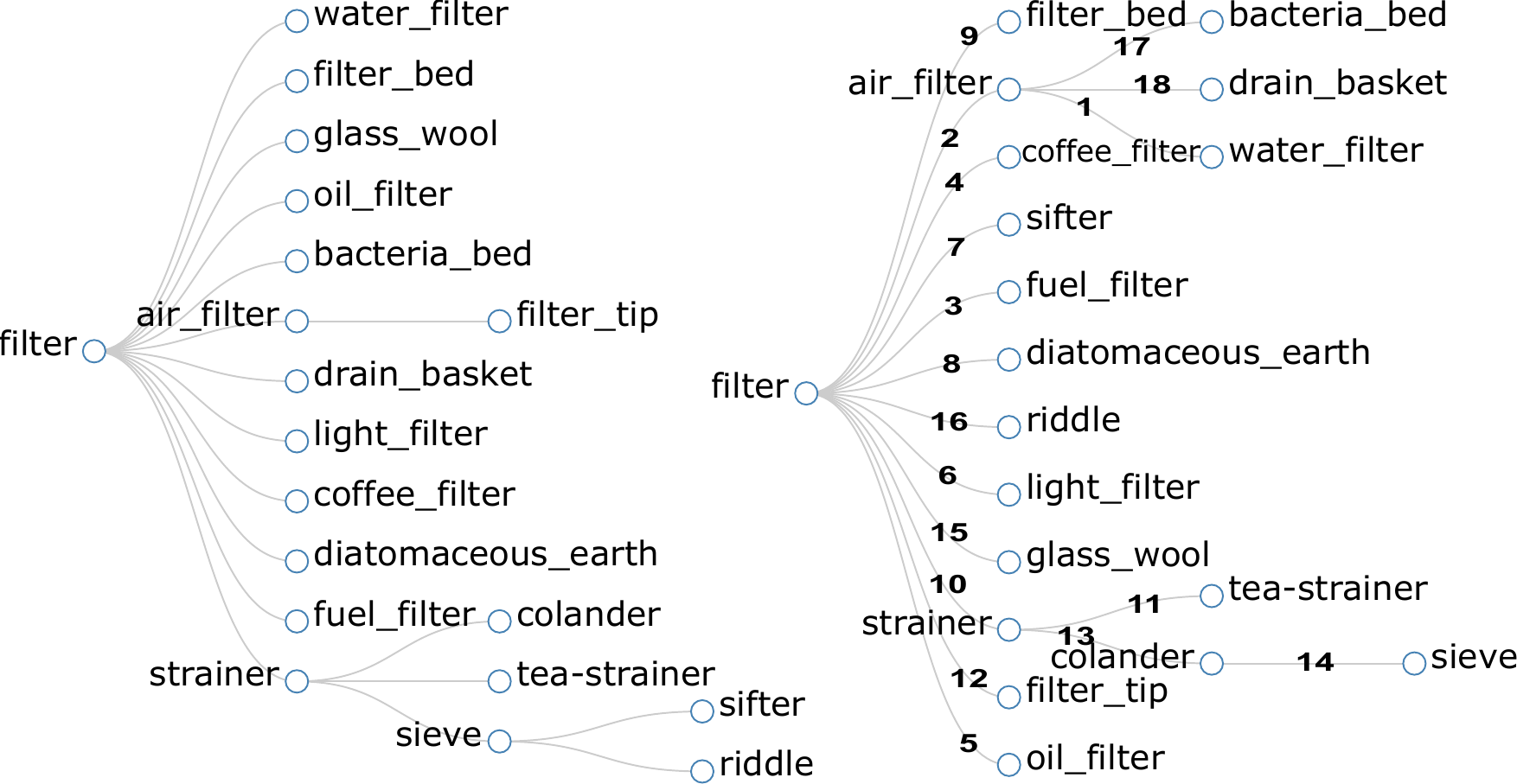}
    \caption{The gold taxonomy in WordNet is on the left. The predicted taxonomy is on the right. The numbers indicate the order of term pair selections. Term pairs with high confidence are selected first.}
    \label{fig:casestudy_filter}
\end{figure}

Fig.~\ref{fig:casestudy_filter} shows the results of taxonomy about \emph{filter}.
We denote the selected term pair at time step $t$ as (hypo, hyper, $t$).
Initially, the term \emph{water filter} is randomly chosen as the taxonomy root. 
Then, a wrong term pair (water filter, air filter, 1) is selected possibly due to the noise and sparsity of features, which makes the term \emph{air filter} become the new root.
(air filter, filter, 2) is selected next and the current root becomes \emph{filter} that is identical to the real root.
After that, term pairs such as (fuel filter, filter, 3), (coffee filter, filter, 4) are selected correctly, mainly because of the substring inclusion intuition.
Other term pairs such as (colander, strainer, 13), (glass wool, filter, 16) are discovered later, largely by the information encoded in the dependency paths and embeddings.
For those undiscovered relations, (filter tip, air filter) has no dependency path in the corpus.
\emph{sifter} is attached to the taxonomy before its hypernym \emph{sieve}.
There is no co-occurrence between \emph{bacteria bed} (or \emph{drain basket}) and other terms. 
In addition, it is hard to utilize the surface features since they ``look different'' from other terms.
That is also why (bacteria bed, air filter, 17) and (drain basket, air filter, 18) are attached in the end: our approach prefers to select term pairs with high confidence first.

\begin{table}[!t]
    \centering
    \small
    \setlength\tabcolsep{5.3pt}
    {\begin{tabular}{|c|c|c|c||c|}
        \hline
        Model & $P_a$ & $R_a$ & $F1_a$ & $F1_e$ \\
        \hline
        \textbf{D}istributional  Info & 27.1 & 24.3 & 25.6 & 13.8 \\
        \hline
        \textbf{P}ath-based  Info & 27.8 & 48.5 & 33.7 & 27.4 \\
        \hline
        \textbf{D} + \textbf{P} & 36.6 & 39.4 & 37.9 & 28.3 \\
        \hline
        \textbf{D} + \textbf{P} + \textbf{S}urface Features & 41.3 & 49.2 & 44.9 & 35.6 \\
        \hline
        \textbf{D} + \textbf{P} + \textbf{S} + FG & \textbf{52.9} & \textbf{58.6} & \textbf{55.6} & \textbf{43.8} \\
        \hline
        \end{tabular}}
    \caption{Ablation study on the WordNet dataset~\cite{bansal2014structured}. $P_e$ and $R_e$ are omitted because they are the same as $F1_e$ for each model. We can see that our approach benefits from multiple sources of information which are complementary to each other.} \label{table_ab}
\end{table}

\section{Related Work} \label{related-work}
\subsection{Hypernymy Detection}
Finding high-quality hypernyms is of great importance since it serves as the first step of taxonomy induction.
In previous works, there are mainly two categories of approaches for hypernymy detection, namely pattern-based and distributional methods.
Pattern-based methods consider lexico-syntactic patterns between the joint occurrences of term pairs for hypernymy detection.
They generally achieve high precision but suffer from low recall.
Typical methods that leverage patterns for hypernym extraction include~\cite{hearst1992automatic,snow2005learning,kozareva2010semi,panchenko2016taxi,nakashole2012patty}.
Distributional methods leverage the contexts of each term separately.
The co-occurrence of term pairs is hence unnecessary.
Some distributional methods are developed in an unsupervised manner. 
Measures such as symmetric similarity~\cite{lin1998information} and those based on distributional inclusion hypothesis~\cite{weeds2004characterising,chang2017unsupervised} were proposed.
Supervised methods, on the other hand, usually have better performance than unsupervised methods for hypernymy detection.
Recent works towards this direction include~\cite{fu2014learning,rimell2014distributional,yu2015learning,tuan2016learning,shwartz2016improving}.

\subsection{Taxonomy Induction}
There are many lines of work for taxonomy induction in the prior literature.
One line of works~\cite{snow2005learning,yang2009metric,shen2012graph,jurgens2015reserating} aims to complete existing taxonomies by attaching new terms in an incremental way.
\citet{snow2005learning} enrich WordNet by maximizing the probability of an extended taxonomy given evidence of relations from text corpora.
\citet{shen2012graph} determine whether an entity is on the taxonomy and either attach it to the right category or link it to an existing one based on the results.
Another line of works~\cite{suchanek2007yagoF,ponzetto2008wikitaxonomy,flati2014two} focuses on the taxonomy induction of existing encyclopedias (\eg, Wikipedia), mainly by employing the nature that they are already organized into semi-structured data.
To deal with the issue of incomplete coverage, some works~\cite{liu2012automatic,dong2014knowledge,panchenko2016taxi,kozareva2010semi} utilize data from domain-specific resources or the Web. 
\citet{panchenko2016taxi} extract hypernyms by patterns from general purpose corpora and domain-specific corpora bootstrapped from the input vocabulary.
\citet{kozareva2010semi}  harvest new terms from the Web by employing Hearst-like lexico-syntactic patterns and validate the learned is-a relations by a web-based concept positioning procedure.


Many works~\cite{kozareva2010semi,anh2014taxonomy,velardi2013ontolearn,bansal2014structured,zhang2016learning,panchenko2016taxi,gupta2017taxonomy} cast the task of hypernymy organization as a graph optimization problem.
\citet{kozareva2010semi} begin with a set of root terms and leaf terms and aim to generate intermediate terms by deriving the longest path from the root to leaf in a noisy hypernym graph.
\citet{velardi2013ontolearn} induct a taxonomy from the hypernym graph via optimal branching and a weighting policy.
\citet{bansal2014structured} regard the induction of a taxonomy as a structured learning problem by building a factor graph to model the relations between edges and siblings, and output the MST found by the Chu-Liu/Edmond's algorithm~\cite{chu1965shortest}.
\citet{zhang2016learning} propose a probabilistic Bayesian model which incorporates visual features (images) in addition to text features (words) to improve the performance. 
The optimal taxonomy is also found by the MST.
\citet{gupta2017taxonomy} extract hypernym subsequences based on hypernym pairs, and regard the task of taxonomy induction as an instance of the minimum-cost flow problem.


\section{Conclusion and Future Work} \label{conclusion}
This paper presents a novel end-to-end reinforcement learning approach for automatic taxonomy induction.
Unlike previous two-phase methods that treat term pairs independently or equally, our approach learns the representations of term pairs by optimizing a holistic tree metric over the training taxonomies.
The error propagation between two phases is thus effectively reduced and the global taxonomy structure is better captured.
Experiments on two public datasets from different domains show that our approach outperforms state-of-the-art methods significantly.
In the future, we will explore more strategies towards term pair selection (\eg, allow the RL agent to remove terms from the taxonomy) and reward function design. 
In addition, study on how to effectively encode induction history will be interesting.


 \section*{Acknowledgments}
 Research was sponsored in part by U.S. Army Research Lab. under Cooperative Agreement No. W911NF-09-2-0053 (NSCTA), DARPA under Agreement No. W911NF-17-C-0099, National Science Foundation IIS 16-18481, IIS 17-04532, and IIS-17-41317, and grant 1U54GM-114838 awarded by NIGMS through funds provided by the trans-NIH Big Data to Knowledge (BD2K) initiative (www.bd2k.nih.gov). The views and conclusions contained in this document are those of the author(s) and should not be interpreted as representing the official policies of the U.S. Army Research Laboratory or the U.S. Government. The U.S. Government is authorized to reproduce and distribute reprints for Government purposes notwithstanding any copyright notation hereon.
 We thank Mohit Bansal, Hao Zhang, and anonymous reviewers for valuable feedback.

\bibliographystyle{acl_natbib}
\bibliography{RL-tree}
\end{document}